\documentclass[
]{ceurart}

\usepackage{listings}
\usepackage{float}
\usepackage{comment}
\usepackage{multirow}
\usepackage{booktabs}
\usepackage{hologo} 
\usepackage{booktabs} 
\usepackage{tablefootnote}
\usepackage{multirow}
\usepackage{xcolor}
\usepackage{graphicx}
\definecolor{mintbg}{HTML}{c5f5e6}
\definecolor{mintframe}{HTML}{9ecbc4}

\usepackage{svg}
\usepackage{hyperref}

\definecolor{olivebg}{HTML}{f1faee}
\definecolor{oliveframe}{HTML}{b0c4b1}

\definecolor{midgreenbg}{HTML}{E2CFB2}
\definecolor{midgreenframe}{HTML}{0B3C49}
\usepackage{subcaption}

\usepackage{hyperref}
\usepackage[most]{tcolorbox}

\lstnewenvironment{codebox}[1][]%
{\lstset{#1}\noindent\minipage{\linewidth}}%
{\endminipage}

\begin{document}

\copyrightyear{2025}
\copyrightclause{Copyright for this paper by its authors.
  Use permitted under Creative Commons License Attribution 4.0
  International (CC BY 4.0).}

\conference{CLEF 2025: Conference and Labs of the Evaluation Forum, September 09–12, 2025, Madrid, Spain}

\title{Harnessing Collective Intelligence of LLMs for Robust Biomedical QA: A Multi-Model Approach}




\author[1,2]{Dimitra Panou}[
orcid=0000-0002-9824-4489,
email=panou@fleming.gr
]
\author[2,3]{Alexandros C. Dimopoulos}[ 
orcid=0000-0002-4602-2040,
email=dimopoulos@fleming.gr,
url=https://dsit.di.uoa.gr/dimopoulos-cv/
]
\author[1,4]{Manolis Koubarakis}[ 
orcid=0000-0002-1954-8338,
email=koubarak@di.uoa.gr,
url=https://cgi.di.uoa.gr/~koubarak/
]

\author[2]{Martin Reczko}[ 
orcid=0000-0002-0005-8718,
email=reczko@fleming.gr,
url=https://www.fleming.gr/research/ifbr/staff-scientists/reczko-lab,
]
\cormark[1]
\address[1]{Department of Informatics and Telecommunications, National and Kapodistrian University of Athens, Greece
}

\address[2]{Institute for Fundamental Biomedical Science,
Biomedical Sciences Research Center "Alexander Fleming", Greece}

\address[3]{Department of Informatics \& Telematics, School of Digital Technology, Harokopio University, Greece}
\address[4]{Archimedes, Athena Research Center, Greece}

\cortext[1]{Corresponding author.}

\begin{abstract}
Biomedical text mining and question-answering are essential yet highly demanding tasks, particularly in the face of the exponential growth of biomedical literature. 
In this work, we present our participation in the 13th edition of the BioASQ challenge, which involves biomedical semantic question-answering for Task 13b and biomedical question-answering for developing topics for the Synergy task. We deploy a selection of open-source large language models (LLMs) as retrieval-augmented generators to answer biomedical questions. Various models are used to process the questions. A majority voting system combines their output to determine the final answer for Yes/No questions, while for list and factoid type questions, the union of their answers in used.
We evaluated 13 state-of-the-art open source LLMs, 
exploring all possible model combinations to contribute to the final answer, 
resulting in tailored LLM pipelines for each question type. 
Our findings provide valuable insight into which combinations of LLMs consistently produce superior results for specific question types. 
In the four rounds of the 2025 BioASQ challenge, our system achieved notable results: in the Synergy task, we secured 1st place for ideal answers and 2nd place for exact answers in round 2, as well as two shared 1st places for exact answers in rounds 3 and 4.

\end{abstract}

\makeatletter\def\Hy@Warning#1{}\makeatother
\maketitle

\section{Introduction}
Large Language Models (LLMs) are transforming numerous fields, but their development and application face distinct challenges tied to accessibility and capabilities.
Closed-source models, such as GPT-models \cite{openai2024gpt4technicalreport}, 
often maintained by large corporations, demonstrate advanced capabilities but lack public accessibility and transparency. 
 In contrast, 
open-source LLMs, grant accessibility, transparency and facilitate fine-tuning and integration into customizable pipelines.

Despite the remarkable progress in LLMs, which are increasingly capable of tackling complex tasks, biomedical QA remains a uniquely challenging field. Effective QA systems must not only retrieve relevant information handling domain-specific terminology, but also discern when to recommend a single "best" option and when to present multiple perspectives.
Avoiding mistakes is critical in this field, as decisions often have direct consequences on human health. Reliable question-answering systems must support experts in exploring these critical issues with accuracy and depth. In fast evolving fields such as drug discovery and molecular biology, where new findings appear constantly and may contradict earlier work, robust tools help professionals stay informed, avoid errors, and make evidence-based decisions that truly advance science and healthcare.

\subsection{The BioASQ Challenge}

The BioASQ challenge has played a central role in advancing biomedical question answering (QA), particularly through its tasks. Task B requires systems to retrieve relevant documents and snippets and then generate precise answers to biomedical questions, while the Synergy track adds further complexity by introducing an interactive, feedback-based QA setting, simulating real-world clinical scenarios. These tasks push the limits of current Retrieval-Augmented Generation (RAG) systems, demanding high precision in both information retrieval and generation. 

Although large language models (LLMs) are becoming increasingly efficient today, the BioASQ challenge demonstrates that achieving accurate results relies on well-structured and carefully designed QA pipelines. Effective systems use hybrid retrievers \cite{mandikal2024sparsemeetsdensehybrid}, domain-specific encoders \cite{biobert2019, pubmedbert}, and fine-tuned generators tailored for biomedical text. Pipelines often include re-ranking steps, prompt tuning, and targeted post-processing to handle subtasks like yes/no classification or list generation. These components are critical to ensure relevance, factuality and clarity, something end-to-end LLMs still struggle with in complex domains. 


Our lab has participated in BioASQ Challenge for three consecutive years. During this period, we experimented with various methodologies to enhance the document selection task. We began by developing our own model, ELECTROLBERT \cite{reczko2022electrolbert}, and later fine-tuned a GAN \cite{panou2023semi} combined with sparse BM25 for document ranking. In our most recent iteration, we transitioned to leveraging existing models for document retrieval, systematically exploring and comparing sparse, dense, and hybrid approaches \cite{panou2024farming}.

Despite improvements in our pipelines, we observed that the Mean Average Precision (MAP) in Phase A remains relatively low for all participants. 
This is mainly because selecting the right documents has become more difficult as the document collection keeps growing. Matching the retrieved documents with the small set chosen by experts remains a challenge. On the other hand, there is still room to improve how answers are generated from the retrieved documents. That’s why in this work, we focus on improving the generation of ‘ideal’ and ‘exact’ answers in Phase B.
\section{Methodology}
\subsection{Synergy}
For the Synergy challenge, we used the same methods as for the final submissions of the BioASQ12 competition, with the notable addition of a DeepSeek-R1 model variant for the generation of exact and ideal answers \cite{DeepSeek-R1-Distill-Llama-70B}. The improved language generation skills of this model led to a notable improvement in the free text required for the ideal answers.

\subsection{Task 13b, phase A: Document Retrieval \& Snippet Identification}
In Phase A and Aplus of the BioASQ challenge, the organizers release biomedical questions curated by experts \cite{nentidis2024,nentidis2025bioasq} that have to be processed within a strict 24-hour interval. For Phase A participants have to retrieve and submit up to 10 relevant documents per question, utilizing abstracts sourced from the PubMed\footnote{\url{https://pubmed.ncbi.nlm.nih.gov/}} database. Based on the retrieved documents, participants must then identify and extract the most relevant snippets.

For document retrieval in Phase A, we adopted a standard approach shown to deliver strong performance in previous work \cite{panou2024farming}, specifically using the BM25 \cite{BM25_2009,ROBERTSON200095} ranking algorithm enhanced with pseudo-relevance feedback from RM3 \cite{RM3}. From this setup, we initially retrieved the top 50 candidate documents and subsequently re-ranked them based on the relevance of their associated snippets. Snippet prediction, which extracts the most semantically relevant snippet from each of the top 10 retrieved documents, is performed as described in \cite{panou2024farming}.




\subsection{Task13b, phase A+ / phase B: exact answer generation}
In Phase A+ participants will submit exact and/or ideal answers before the expert selected (gold) documents and snippets (released in Phase B) are known.  It serves as a baseline to compare with Phase B, where feedback is provided to guide system improvement.
Each participant must rely on their own predicted documents and snippets for subsequent processing. They have 24 hours to submit their results, which include documents, snippets, exact and "ideal" answers, based on the provided test set. For document selection, we followed the same procedure as in Phase A. To generate the exact and "ideal" answers, we used both the predicted snippets and the full abstracts as input.

\begin{figure}[htp]
    \centering
    \includegraphics[width=16cm]{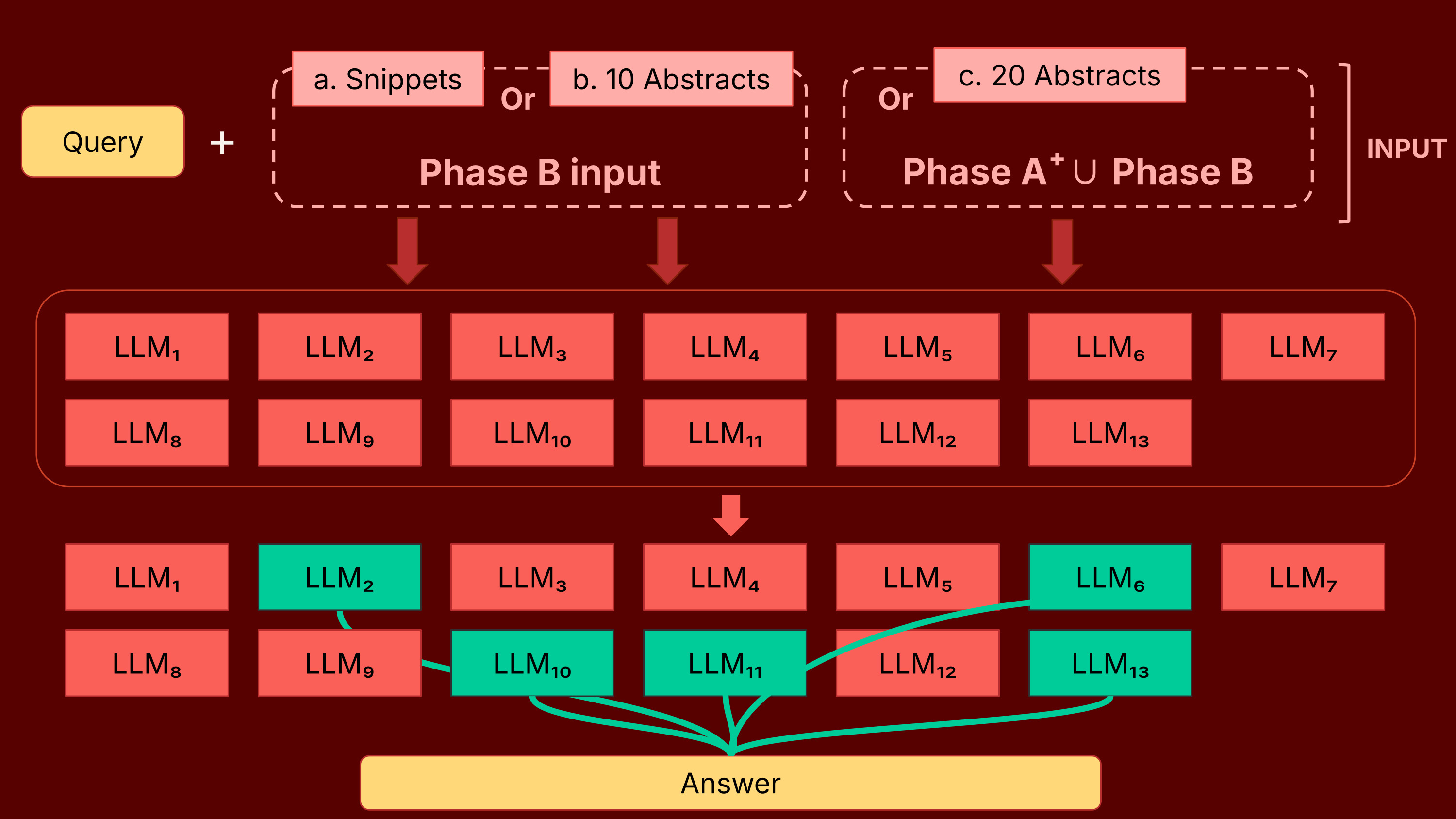}
    \caption{\footnotesize{Processing during phase A+ and phase B. a. Prediction based on given snippets, b. Prediction based on given abstracts, c. Prediction based on given and predicted abstracts. The query and one of the three input alternatives are used to form the prompts for the LLMs. The collection of all LLMs used to find optimal sets based on the training set is shown in the middle, the identified optimal subset used for prediction is shown in green at the bottom. } } 

    \label{multiple_models}
\end{figure}

In Phase B, participants are required to submit exact answers for Yes/No, List, and Factoid questions, as well as ideal answers for summary-type questions. This phase uses gold-standard documents and snippets. In Phase B, 
we explored three distinct approaches for generating exact answers, as illustrated in Figure\ref{multiple_models}.

The first approach (Figure \ref{multiple_models}a) utilizes the extracted snippets from the given (golden) documents, incorporating them directly into the prompt to generate answers for each question. This method has been used in our previous submissions and is commonly adopted by participants in the BioASQ challenge. It is computationally efficient, as the snippets are typically short, ranging from a few words to two sentences.
The second approach (Figure~\ref{multiple_models}b) uses the full abstracts of the top 10 most relevant documents. The prompt is constructed by combining the question with these abstracts in the following format: \texttt{text = <Abstract 1>, <Abstract 2>, ..., <Abstract 10>}, as shown in Appendix \ref{Appendix}. This method provides broader contextual information and outperformed the first approach in our evaluations.

The third approach (Figure~\ref{multiple_models}c) builds upon the second by additionally incorporating any relevant documents identified during the extended document retrieval process in Phase A+. These supplementary documents are appended to the original list, further enriching the input context provided to the model and potentially including documents not selected by the experts who created the gold answers.

\begin{table}[h!]
\centering
\begin{tabular}{cllll}
\hline
\# & Abbreviation     & Model Name                            & Edition / Quantization     & Parameter Size \\
\hline
1  & Reflection       & Reflection\footnotemark[1]                             & latest                     &70B       \\
2  & L3.1         & LLaMA 3.1\footnotemark[2]         & Q4 / ctx8192                & 70B            \\
3  & L3.3         & LLaMA 3.3\footnotemark[3]                              & latest                     & 70B            \\
4  & Mixtral          & Mixtral\footnotemark[4]                              & ctx8192:latest               & 8x7B           \\
5  & Qwen14          & Qwen3:14B\footnotemark[6]                  & latest                     & 14B             \\
6  & Qwen30     & Qwen3:30B-A3B\footnotemark[7]                           & A3B                         & 30B             \\
7  & Qwen32          & Qwen3:32B\footnotemark[8]                              & latest                     & 32B             \\
8  & Yi               & Yi\footnotemark[8]                                     & latest                         & 34B             \\
9 & Smaug         &Smaug:72B\footnotemark[9]              &Q4\_K\_M,  quantized 4 bit                   & 72B             \\
10 & DSQ8             & DeepSeek-R1-Distill-Llama-70B\footnotemark[10]\cite{DeepSeek-R1-Distill-Llama-70B}& Q8\_0:latest, quantized 8 bit                         &70B              \\
11 & Phi3      & Phi-3 Medium\footnotemark[11]                           & latest                      & 14B           \\

12 & Phi4            & Phi-4\footnotemark[12]                                  & latest                     & 14B         \\
13 & Aya           & Aya:35B\footnotemark[13]                                & latest                     & 35B             \\
\hline
\end{tabular}
\caption{Large language models used individually and in all combinations of them.}
\label{tab:tested_models}
\end{table}
\footnotetext[1]{\url{https://huggingface.co/mattshumer/ref_70_e3}}
\footnotetext[2]{\url{https://ollama.com/library/llama3.1:70b}}
\footnotetext[3]{\url{https://huggingface.co/meta-llama/Llama-3.3-70B-Instruct}}
\footnotetext[4]{\url{https://huggingface.co/mistralai/Mixtral-8x7B-Instruct-v0.1}}
\footnotetext[5]{\url{https://huggingface.co/Qwen/Qwen3-14B}}
\footnotetext[6]{\url{https://huggingface.co/Qwen/Qwen3-30B-A3B}}
\footnotetext[7]{\url{https://huggingface.co/Qwen/Qwen3-32B}}
\footnotetext[8]{\url{https://huggingface.co/01-ai/Yi-34B}}
\footnotetext[9]{\url{https://huggingface.co/senseable/Smaug-72B-v0.1-gguf/blob/main/Smaug-72B-v0.1-q4_k_m.gguf}}
\footnotetext[10]{\url{https://huggingface.co/unsloth/DeepSeek-R1-Distill-Llama-70B-GGUF}}
\footnotetext[11]{\url{https://ollama.com/library/phi3:medium}}
\footnotetext[12]{\url{https://ollama.com/library/phi4}}
\footnotetext[13]{\url{https://ollama.com/library/aya:35b}}


To improve the answer performance measures, we employed an LLM "farming" strategy, which we initially implemented last year for Yes/No questions. This strategy utilizes a diverse ensemble of complementary open-source large language models. In the present study, we extend this strategy to all exact answer types, aggregating the union of answers from multiple LLMs for factoid and list questions.

Using the BioASQ11 and BioASQ12 training set, we evaluated 13 state-of-the-art LLMs using Ollama \cite{ollama} and LM Studio \cite{lightningstudio2023}, systematically analyzing their individual performance, as well as all possible combinations of models for each type of question. This experimentation allowed us to construct an optimal ‘farm’ of models for each category of question. Due to the long run-time of these optimizations, our submissions in the competition did not represent the finally best performing system. For all types of questions, the optimization revealed novel combinations of models with higher performance than any single LLM.

\subsubsection{Optimal factoid question answering subsets}

\begin{figure}[htp]
    \centering
    \includegraphics[width=16cm]{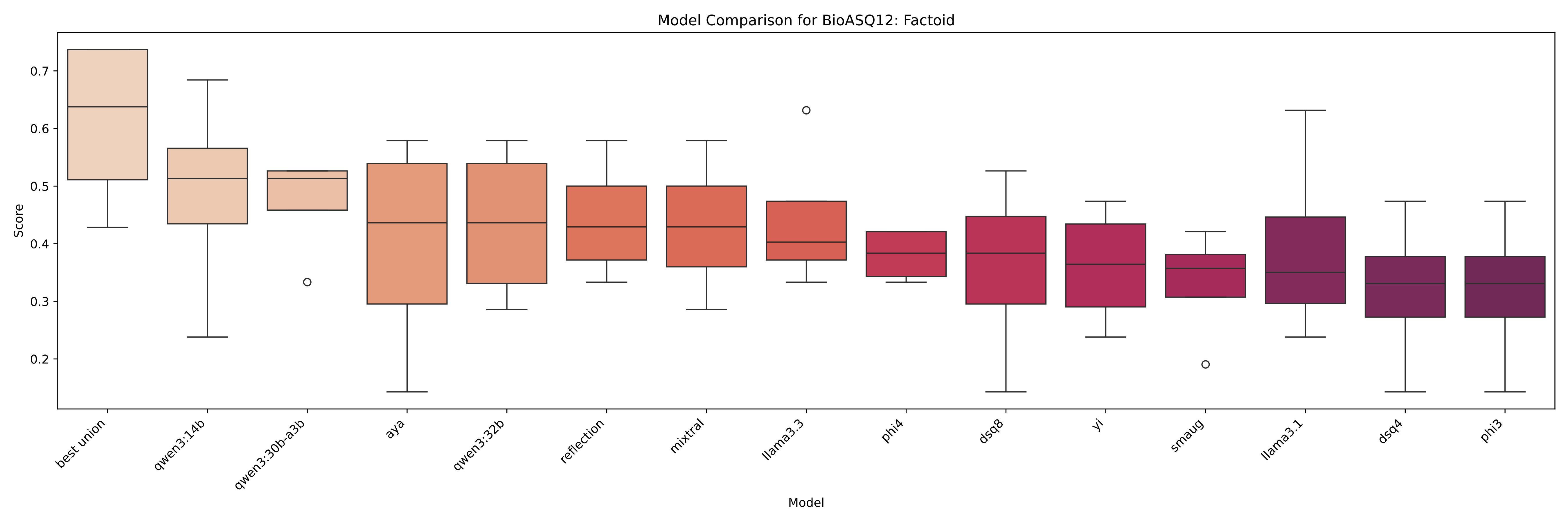}
    \caption{\footnotesize{Comparison of single LLMs and the optimal union of LLMs for factoid questions performance tested on BioASQ12 datasets.}}
    \label{factoid_model_comparison}
\end{figure}

\begin{figure}[htp]
    \centering
    \includegraphics[scale =0.15]{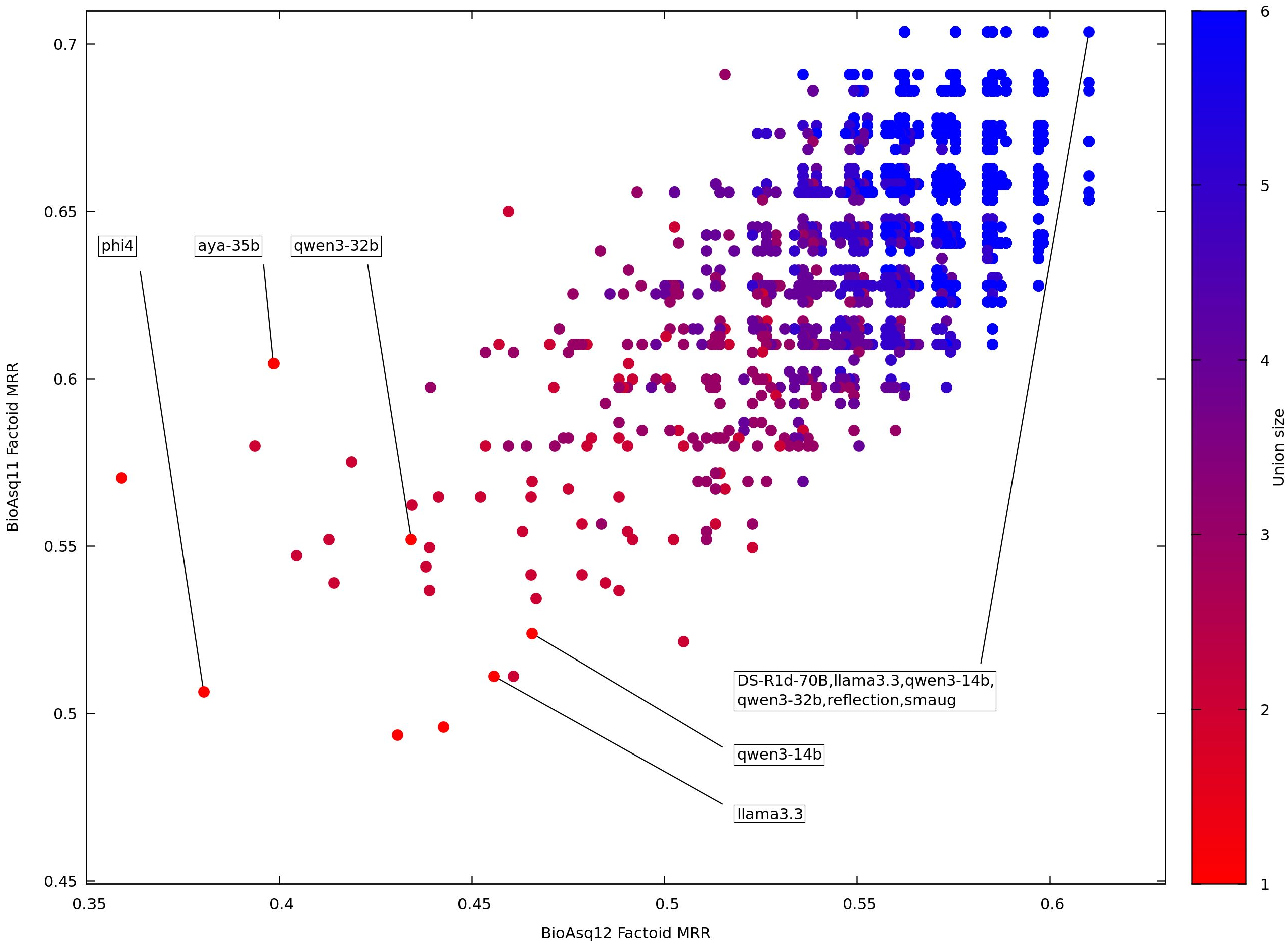}
    \caption{\footnotesize{Model unions for factoid questions performance tested on BioASQ11 \& BioASQ12 datasets} } 
    \label{factoid_models_png}
\end{figure}

For the 13 LLMs listed in table \ref{tab:tested_models}, there are $| \{ S_i , i \in [1,2^{13}] \}| = 8191$ different subsets. The 13 LLMs predict sets of factoids for all questions in the BioASQ11 and BioASQ12 training sets, and the factoid sets for each LLM in $S_i$ are combined to form a union for each question. Since the factoids should be ordered by relevance and only the top 5 most relevant should be returned, the combination of the factoids considers the relevance scores that are also returned by each LLM. The performance of these sets is evaluated with the usual Mean Reciprocal Rank ($MRR$) measure. The $MRR$ values, averaged over four rounds, for each $S_i$ are shown in the scatterplot \ref{factoid_models_png} that visualizes the performances in BioASQ11 and BioASQ12.
The color of each dot indicates the size of the LLM set. Single LLMs are shown in red, the largest sets containing 6 LLMs are shown in blue. It should be noted that in all cases the sets with more than 6 LLMs had the same performance as a "kernel" set of 6 LLMs and are not contained in the plot. It can be clearly seen that all single LLMs had worse performances than any union of at least 4 LLMs (see also figure \ref{factoid_model_comparison}). Consistently, the highest performances are obtained with unions of 6 LLMs.
These observations can be made for both BioASQ11 and BioASQ12 independently, indicating no training set specificities. The finding that larger unions give better results is very likely due to the complementarity of the answers of the different LLMs. There can be many cases where one method finds a highly relevant factoid that another method does not identify at all or as a close miss. The merging strategy that uses the confidence scores supports these situations.

\begin{figure}[htp]
    \centering
        \centering
        \includegraphics[width=0.90\textwidth]{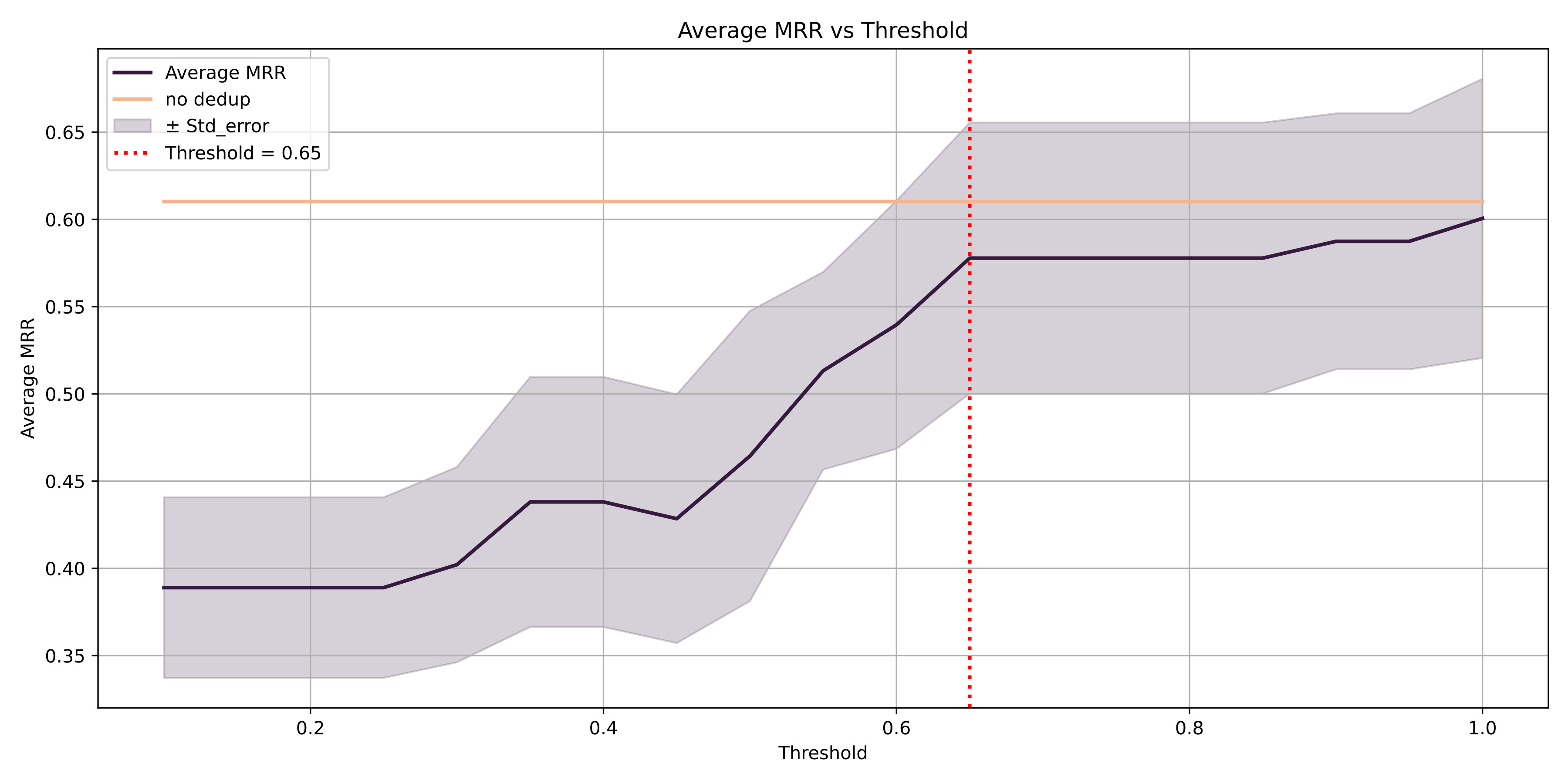}
        \caption{\footnotesize Deduplication performance for factoid questions}
        \label{factoid_dedup}
\end{figure}

\paragraph{Factoid deduplication}
As forming the union might introduce multiple occurrences of exactly the same factoid phrase or semantically similar phrases, we investigated a simple deduplication procedure. Each factoid phrase is embedded with a standard transformer (all-MiniLM-L6-v2) and the cosine similarity of the embedding between all factoids is measured. With different thresholds for the cosine similarity, semantically similar phrases can be removed from the set. The $MRR$ performance with different thresholds for the LLM subset with the best performance on BioASQ12 is shown in figure \ref{factoid_dedup}. It can be observed that deduplication does not improve $MRR$ performance, consistent with our observation that larger subsets in general have higher $MRR$ performances.

\subsubsection{Optimal list question answering subsets}
\begin{figure}[htp]
    \centering
    \includegraphics[width=16cm]{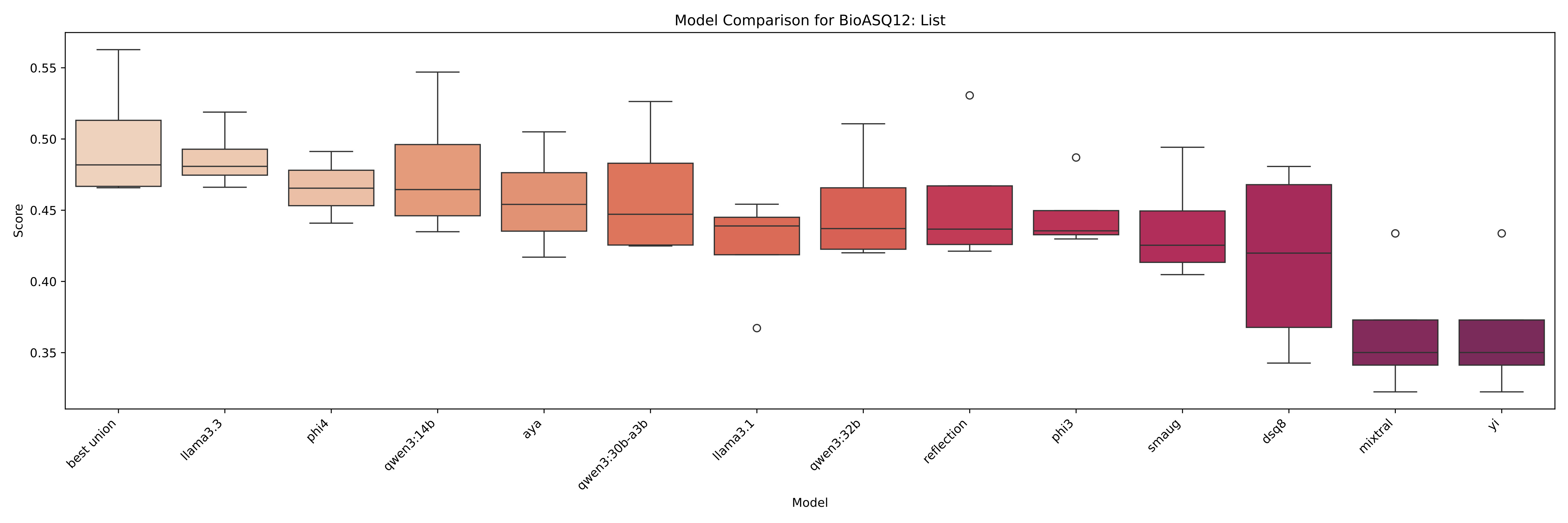}
    \caption{\footnotesize{Comparison of single LLMs and the optimal union of LLMs for list questions performance tested on BioASQ12 datasets.}}
    \label{list_model_comparison}
\end{figure}

\begin{figure}[htp]
    \centering
    \includegraphics[width = 0.80\textwidth]{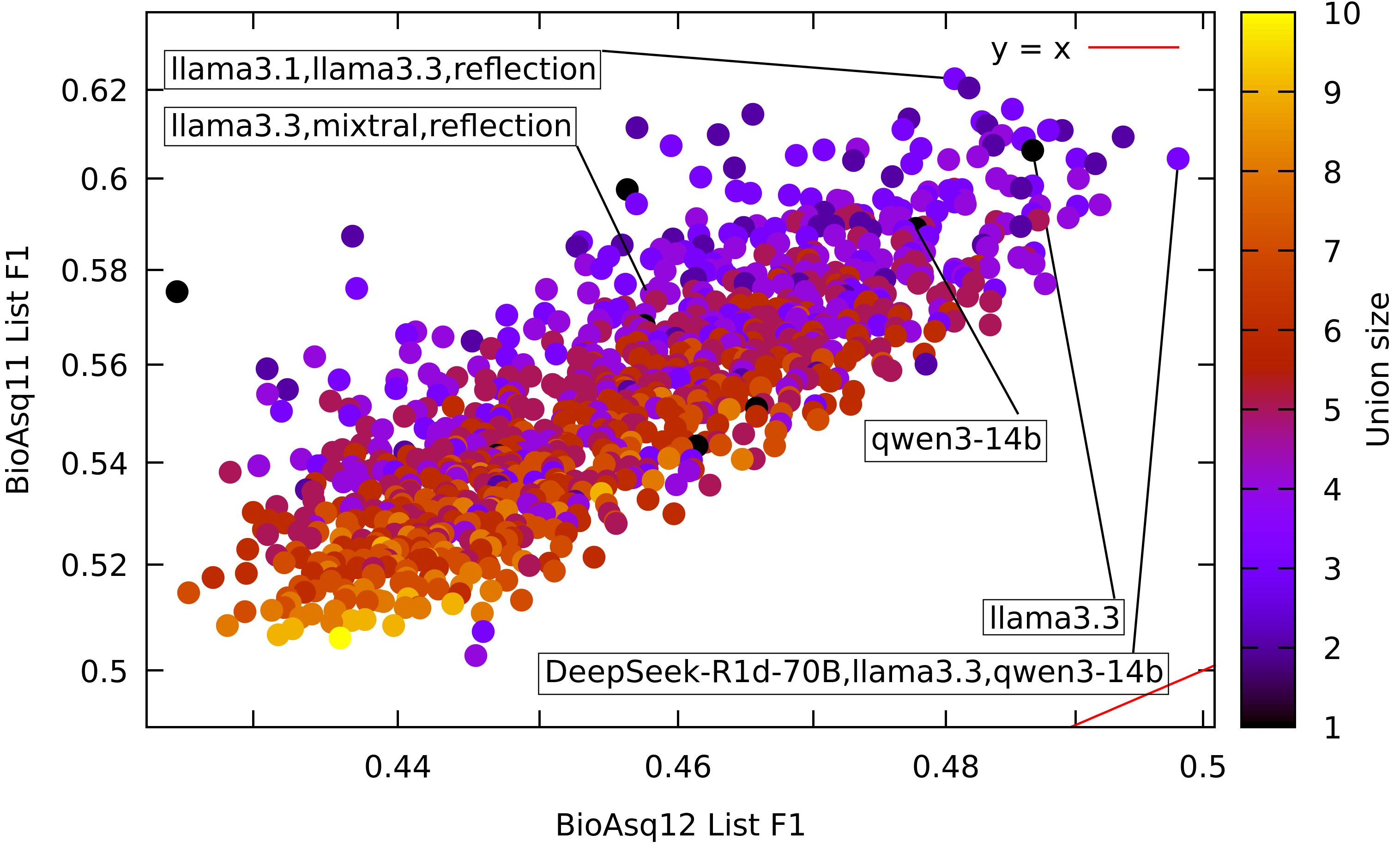}
    \caption{\footnotesize{Model unions for List questions performance tested on BioASQ11 \& BioASQ12 datasets} }
    \label{list_models_png}
\end{figure}

A procedure similar to the processing of the factoid questions is used for list type questions. As no relevance order and no limit is required for the list items in the answer, the set of list items is the simple union of the list items predicted by each LLM in the subset $S_i$. The usual performance measure for list type questions is F1, which is the harmonic mean of precision and recall. With a growing size of the list items in the union for each additional LLM in the subset $S_i$, the chance of false positive items increases and precision decreases. In the scatterplot showing the F1 scores (averaged over four rounds) for BioASQ11 and BioAQS12 in figure \ref{list_models_png}, it can be clearly seen that the large subsets with more than 7 LLMs have significantly lower performance than the smaller subsets, independently of the BioASQ dataset used. However, as detailed in figure \ref{list_model_comparison}, several specific combinations, such as the set [DSQ8, L3.3, Qwen14] have better performance than any of the single LLMs.

\begin{figure}[htp]        
        \centering
        \includegraphics[width=0.90\textwidth]{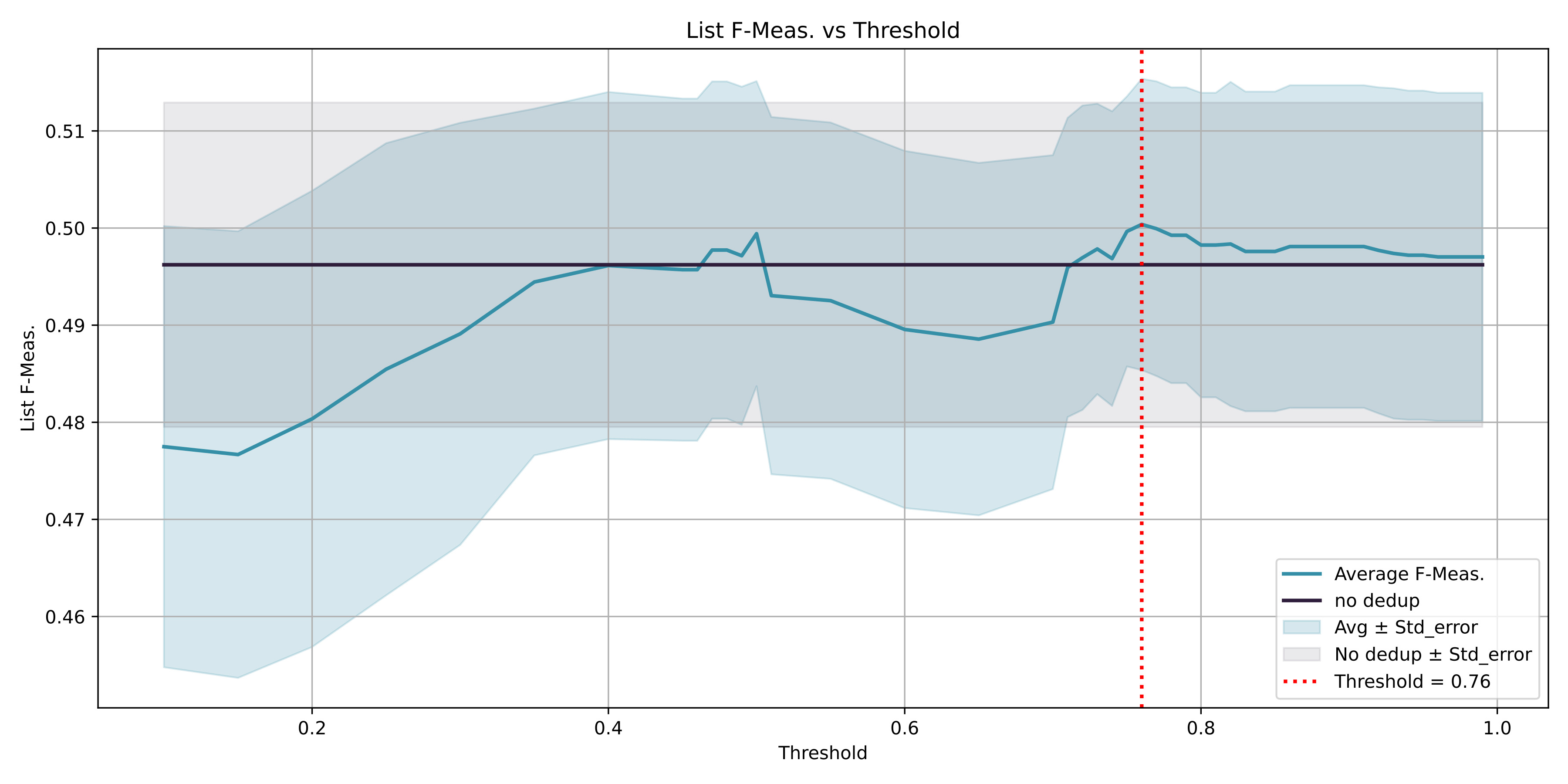}
        \caption{\footnotesize Deduplication performance for list questions}
        \label{list_dedup}
\end{figure}

\paragraph{List deduplication}
The same deduplication procedure used for factoids is also evaluated for the union of list items. For the subset with the best performance on the BioASQ12 set, the F1 performance for different thresholds for the cosine similarity is shown in figure \ref{list_dedup}. It can be observed that two levels of deduplication achieve higher performance than without deduplication, with an optimal F1 values at a threshold of $0.76$. A threshold of $0.7$ was used for the list type submission in the BioASQ13 competition.

\subsubsection{Optimal jury sets answering Yes/No questions}
\begin{figure}[htp]
    \centering
    \includegraphics[scale =0.12]{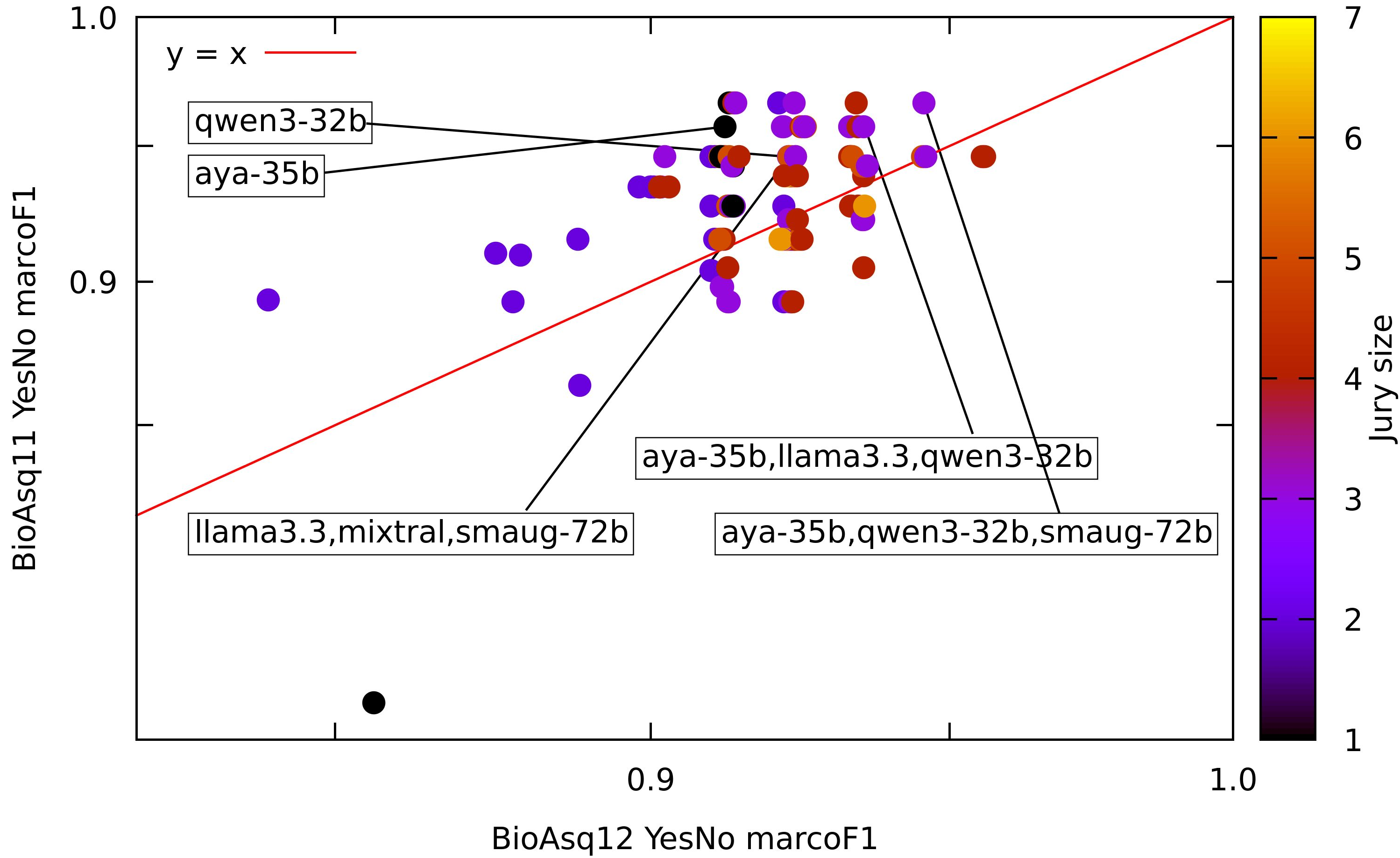}
    \caption{\footnotesize{Model juries for Yes/No questions performance tested on BioASQ11 \& BioASQ12 datasets} }
    \label{yesno_models_png}
\end{figure}

The concept of using a jury (or  ‘farm’) of LLMs was introduced by our lab for BioASQ12 \cite{panou2024farming}. Here we further optimize this by evaluating all possible combinations of LLMs and adding more recent LLMs. A subset $S_i$ of LLMs generates an answer by counting the number of "Yes" and "No" outcomes for each participating LLM. The final answer will be "Yes" if there are a higher or equal number of "Yes" outcomes than "No" outcomes. The performances of the different subsets with the usual macroF1 measure (averaged over four rounds) is shown for BioASQ11 and BioASQ12 in figure \ref{yesno_models_png}. The discrete nature of this question type leads to more discrete performance levels that are visualized in the plot by applying a small jitter. As in the case of the list type questions, it can be seen that there are several combinations of a few LLMS like [Aya, Qwen32, Smaug] that outperform any of the individual LLM alone.

\begin{figure}[htp]
    \centering
    \includegraphics[width=16cm]{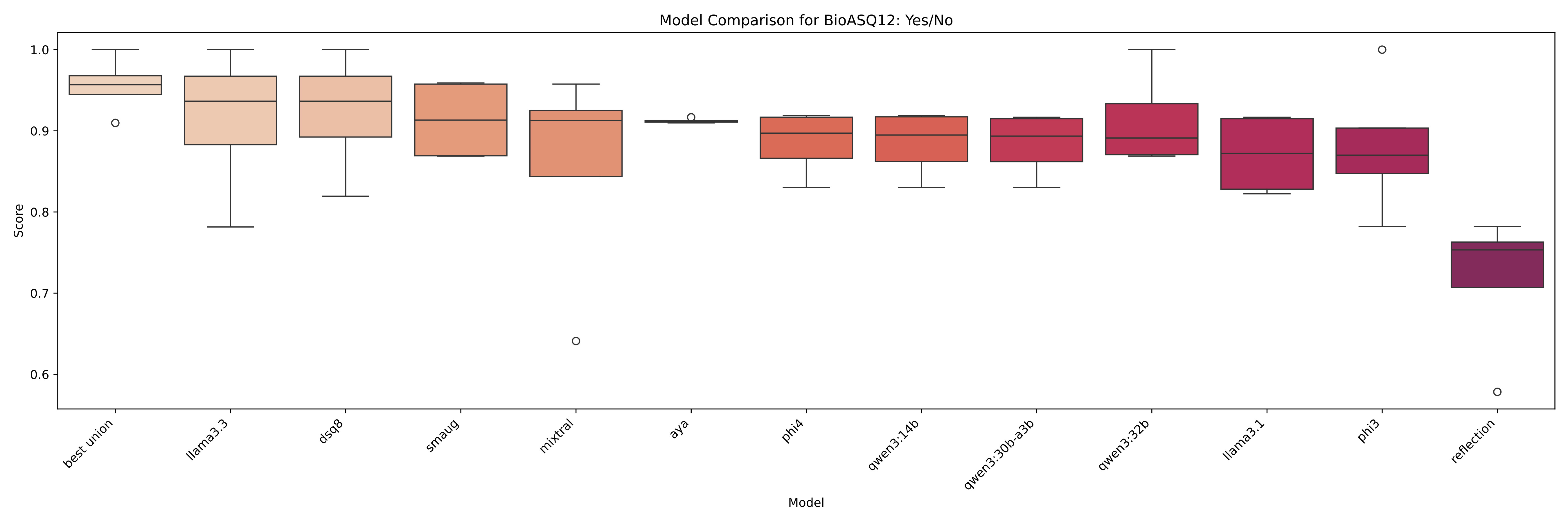}
    \caption{\footnotesize{Comparison of single LLMs and the optimal union of LLMs for Yes/No questions performance tested on BioASQ12 datasets.}}
    \label{yesno_model_comparison}
\end{figure}

\section{Results}
The evaluation of systems participating in the BioASQ competition Task B varies \cite{malakasiotis2018evaluation} based on the question type. 

\subsection{Synergy Results}
In the four rounds of the Synergy 2025 BioASQ challenge, our system achieved notable results: first place in round 2 for "ideal answers" and second place in rounds 3 and 4. 
\\As shown in Table~\ref{tab:Synergy_exact}, the evaluation for "Exact" answers includes multiple measures across three question types: Accuracy and Macro F1 for Yes/No questions, Mean Reciprocal Rank (MRR) and Lenient Accuracy for Factoid questions, and Mean Precision and F-measure for List questions. The overall position per system in each batch is calculated based on a combination of Macro F1 (Yes/No), MRR (Factoid), and F-measure (List). While the top-ranked systems achieve the highest combined scores, the Fleming submissions stand out with stronger results in the List category and competitive performance in Factoid. Table~\ref{tab:Synergy_ideal} presents the performance of systems on the "Ideal answers" task, as assessed through manual evaluation. The scores reflect human judgments across four criteria: Readability, Recall, Precision, and Repetition, with the final Mean Manual score representing their average. In Batch 2, the Fleming system achieved the highest overall score, ranking 1\textsuperscript{st}, while in Batches 3 and 4, it remained competitive with particularly strong Recall and Repetition scores, securing 2\textsuperscript{nd} and 4\textsuperscript{th} place respectively.


\begin{table}[htbp]
\centering
\footnotesize
\caption{Synergy: Exact answers performance measured by the combination of Macro F1, MRR, and F-measure}
\begin{tabular}{cccccccccc}
\toprule
Batch & Position & System & \multicolumn{2}{c}{Yes/No} & \multicolumn{2}{c}{Factoid} & \multicolumn{2}{c}{List} \\
\cmidrule(lr){4-5} \cmidrule(lr){6-7} \cmidrule(lr){8-9}
& & & Macro F1 & Rank & MRR & Rank & F-Measure & Rank \\
\midrule
Batch 2 & 1/10 & \textbf{dmiip2024\_1} & 1.000 & 1 & 0.4286 & 1&  0.2467 & 1 \\
& 2/10 & dmiip2024\_2 & 1.000 & 1 & 0.2857 & 2&  0.2000 & 4 \\
& 3/10 & Fleming-1 & 0.8571	& 2 & 0.2857& 2&  \textbf{0.2100} & 3 \\
& 7/10 & Fleming-1 & 0.5333	& 5 & 0.2857& 2&  \textbf{0.2333} & 2 \\
\hline
Batch 3 & 1/13 & \textbf{dmiip2024\_4} & 0.899 & 2 & 0.5000 & 1 & 0.2495 & 3\\
& 2/13 & Fleming-3 & 0.899 & 2 & 0.2500 & 3 & \textbf{0.2634} & \textbf{1} \\
& 4/13 & Fleming-1 & 0.7917 & 3 & 0.2500 & 3 & \textbf{0.2634} & \textbf{1} \\
& 5/13 & Fleming-2 & 0.7917 & 3 & 0.2500 & 3 & \textbf{0.2634} & \textbf{1} \\
\hline
Batch 4 & 1/15 & \textbf{sinai\_uja\_RAG} & 0.899 & 2 & 0.4000 & 2 & 0.2667 & 2 \\
& 2/15 & \textbf{Fleming-4} & 0.7917 & 3&  0.4000 & 2 & \textbf{0.3536} & 1 \\
& 3/15 & \textbf{Fleming-1} & 0.7917 & 3&  0.4000 & 2 & \textbf{0.3536} & 1 \\
& 4/15 & \textbf{Fleming-2} & 0.7917 & 3&  0.4000 & 2 & \textbf{0.3536} & 1 \\
& 5/15 & \textbf{Fleming-3} & 0.7917 & 3&  0.4000 & 2 & \textbf{0.3536} & 1 \\
\bottomrule
\end{tabular}
\label{tab:Synergy_exact}
\end{table}

\begin{table}[htbp]
\centering
\footnotesize
\caption{Synergy: "Ideal answers" performance measured  by mean of manual score.}
\begin{tabular}{ccccccccc}
\toprule
Batch & Position & System & Readability & Recall & Precision & Repetition & Mean Manual \\
\midrule
Batch 2 & 1/10 & \textbf{Fleming-1} & 4.06&	4.24&	3.88&	4 &  4.045 \\
& 2/10 & dmiip2024\_1 & 4.06 & 3.79	& 3.79	& 4.39 & 4.0075 \\
& 5/10 & Fleming-2 & 3.52 & 4.06	& 3.27	& 3.27 & 3.730 \\
\hline
Batch 3 & 1/13 & \textbf{dmiip2024\_1} & 4.57&	4.61 & 4.43&	4.57 &4.545 \\
& 6/13 & Fleming-1 & 4.31&	\textbf{4.73}&	4.06&	4.47 & 4.3925 \\
& 7/13 & Fleming-2 & 4.31&	\textbf{4.73}&	4.06&	4.47 & 4.3925 \\
& 8/13 & Fleming-3 & 4.31&	\textbf{4.73}&	4.06&	4.47 & 4.3925 \\
\hline
Batch 4 & 1/15 & \textbf{dmiip2024\_1} & 4.47&	4.53& 4.24& 4.49& 4.4325\\
& 9/15 & Fleming-1 & 4.05&	4.51& 3.69& 4.09 & 4.085& \\
& 10/15 & Fleming-3 & 4.05&	4.51& 3.69& 4.09 & 4.085& \\
& 11/15 & Fleming-4 & 3.98&	4.36& 3.56& 4.09 & 3.998& \\
& 12/15 & Fleming-2 & 3.91&	4.51& 3.47& 3.96 & 3.963& \\
\bottomrule
\end{tabular}
\label{tab:Synergy_ideal}
\end{table}

\subsection{Task 13b: Phase A}

\subsubsection{Document retrieval}
In Table \ref{PhaseA_Documents} the preliminary performances of our document retrieval submissions for the BioASQ13 competition are listed. The final and official results, will be available shortly before the BioASQ13 workshop, after the manual assessment of all system responses by the BioASQ experts and the enrichment of the respective ground truth with potential additional correct elements. 

\begin{table}[htbp]
\centering
\caption{Phase A: System performance for Document retrieval measured as mean average precision ($MAP$)}
\begin{tabular}{ccccccccc}
Batch & Position & System & Mean Precision & Recall & F-Measure & MAP & GMAP \\
\toprule
Batch 1 & 1/51 & \textbf{bioinfo-4}   & 0.1047 & 0.5043 & 0.1605 & 0.4246 & 0.0104  \\
       & 24/51 & Fleming-1   & 0.0606 & 0.3863 & 0.1005 & 0.2716 & 0.0020 \\
\midrule
Batch 2 & 1/42 &\textbf{Baseline top 10} & 0.0976 & 0.5093 & 0.1546 & 0.4425 & 0.0096 \\
    & 18/42   & Fleming-2   & \textbf{0.0993} & 0.4333 & 0.1477 & 0.3066 & 0.0026 \\
    & 19/42   & Fleming-3   & \textbf{0.0993} & 0.4333 & 0.1477 & 0.3066 & 0.0026\\
    & 22/42  & Fleming-1   & 0.0861 & 0.4333 & 0.1342 & 0.2957 & 0.0026\\
\midrule
Batch 3 & 1/47 &\textbf{bioinfo-1}   & 0.0941 & 0.4228 & 0.1445 & 0.3236 & 0.0059\\
    & 25/47    & Fleming-1   & 0.0697 & 0.3105 & 0.1064 & 0.1794 & 0.0009\\
\midrule
Batch 4 & 1/79  &\textbf{bioinfo-1} & 0.06 &0.2512& 0.0927&	0.1801&	0.0008\\
    & 24/79   & Fleming-2   & 0.0383& 0.155& 0.05957& 0.09427&  0.0002\\
    & 25/79   & Fleming-1   & 0.0383& 0.155& 0.0595& 0.0863& 0.0002\\
\bottomrule
\end{tabular}
\label{PhaseA_Documents}
\end{table}

\subsection{Task 13b: Phase A+ and Phase B}
\subsubsection{Exact answer prediction}
The tables reporting the Phase A+ (Table \ref{tab:PhaseAplus_exact}) and Phase B (Table \ref{tab:PhaseB_exact}) results of BioASQ13, for exact answers provide a comparative view of our submitted systems. In each batch, the first row corresponds to the top-ranked competitor. 
For each question type, we report a corresponding evaluation metric: Macro F1 for Yes/No, MRR for Factoid, and F-Measure for List. The systems are ranked per metric and the total rank is computed as the sum of these individual ranks, providing an overall measure of performance across all type of questions. The final position according to the total rank and the total number of submissions is indicated in the column "Position". Our systems demonstrated competitive performance, particularly in the Yes/No and Factoid categories. 

\begin{table}[htbp]
\centering
\footnotesize
\caption{Phase A+: Exact answers performance measured by the combination of Macro F1, MRR, and F-measure}
\begin{tabular}{l l l rr rr rr }
\toprule
Batch & Position & System & \multicolumn{2}{c}{Yes/No} & \multicolumn{2}{c}{Factoid} & \multicolumn{2}{c}{List} \\
\cmidrule(lr){4-5} \cmidrule(lr){6-7} \cmidrule(lr){8-9}
& & & Macro F1 & Rank & MRR & Rank & F-Measure & Rank \\
\hline
Batch 1 & 1/56 & \textbf{UR-IW-2} & \textbf{1.000} & 1 & 0.3782 & 5 & 0.2567 & 1 \\
& 20/56 & Fleming-3 & 0.9328 & 2 & 0.3186 & 11 & 0.144 & 26 \\
& 21/56 & Fleming-2 & 0.9328 & 2 & 0.3186 & 11 & 0.144 & 26 \\
& 31/56 & Fleming-1 & 0.9328 & 2 & 0.3186 & 11  & 0.1296 & 33 \\
\hline
Batch 2 & 1/49 & \textbf{Baseline top 20} & 0.9328 & 3 & 0.463 & 6 & 0.388 & 1 \\
& 29/49 & Fleming-1 & \textbf{0.9377}	& \textbf{2} & 0.2790 & 20 & 0.2242 & 25 \\
& 32/49 & Fleming-2 & 0.9328 & 3 & 0.2790 & 20 & 0.2242 & 25 \\
\hline
    Batch 3 & 1/58 & \textbf{IR3} & 0.6944 & 11 & 0.3500 & 2 & 0.4313 & 4 \\
& 6/58 & Fleming-2 & \textbf{0.8182} & \textbf{6} & 0.3125 & 4 & 0.3565 & 15 \\
& 19/58 & Fleming-1 & 0.6563 & 13 & 0.2625 & 11 & 0.3565 & 15 \\
\hline
    Batch 4 & 1/67 & \textbf{Baseline top 20} & 0.8595 & 4 & 0.4318 & 8 & 0.2977 & 2 \\
& 31/67 & Fleming-1 & \textbf{0.8595} & \textbf{4} & 0.2803 & 19 & 0.2425 & 24 \\
& 32/67 & Fleming-4 & \textbf{0.8595} & \textbf{4} & 0.2780 & 20 & 0.2433 & 23 \\
& 53/67 & Fleming-2 & \textbf{0.8595} & \textbf{4} & 0.2818 & 18 & 0.1578 & 50 \\
& 60/37 & Fleming-3 & 0.7068 & 17 & 0.2818 & 18 & 0.1578 & 50 \\
\bottomrule
\end{tabular}
\label{tab:PhaseAplus_exact}
\end{table}

\begin{table}[htbp]
\centering
\footnotesize
\caption{Phase B: Exact answers performance measured by the combination of Macro F1, MRR, and F-Measure}
\begin{tabular}{l l l rr rr rr }
\toprule
Batch & Position & System & \multicolumn{2}{c}{Yes/No} & \multicolumn{2}{c}{Factoid} & \multicolumn{2}{c}{List} \\
\cmidrule(lr){4-5} \cmidrule(lr){6-7} \cmidrule(lr){8-9}
& & & Macro F1 & Rank & MRR & Rank & F-Meas. & Rank \\
\midrule
Batch 1 & 1/72 & \textbf{2025-DMIS-KU-3} & 0.9328 & 2 & \textbf{0.5962} & \textbf{1} & 0.5913 & 2 \\
& 6/72 & Fleming-3 & 0.9244 & 3 & 0.5577 & 2 & 0.5384 & 14 \\
& 13/72 & Fleming-1 & \textbf{1.0000} & \textbf{1} & \textbf{0.5962} & \textbf{1} & 0.5290 & 20 \\
& 15/72 & Fleming-2 & 0.9244 & 3 & 0.5962 & 1 & 0.5290 & 20 \\
\midrule
Batch 2 & 1/72 & \textbf{dmiip2024\_4} & \textbf{1.0000} & \textbf{1} & 0.5926 & 6 & 0.6152 & 1 \\
& 25/72 & Fleming-2 & \textbf{1.0000} & \textbf{1} & 0.4704 & 19 & 0.5356 & 19 \\
& 27/72& Fleming-3 & \textbf{1.0000} & \textbf{1}& 0.5148 & 15 & 0.5210 & 24 \\
& 35/72 & Fleming-1&  \textbf{1.0000} & \textbf{1} & 0.4704 & 19 & 0.5210 & 24 \\
\midrule
Batch 3 & 1/66 & \textbf{EP-1} & \textbf{0.9394} & \textbf{1} & 0.4625 & 4 & 0.6331 & 2 \\
& 27/66 & Fleming-1 & \textbf{0.9394} & \textbf{1} & 0.2717 & 23 & 0.5638 & 22 \\
& 38/66 & Fleming-2 & 0.8706 & 3 & 0.3083 & 19 & 0.4832 & 40 \\
& 47/66 & Fleming-4 & \textbf{0.9394} & \textbf{1} & 0.3225 & 18 & 0.4595 & 48 \\
& 49/66 & Fleming-3 & \textbf{0.9394} & \textbf{1} & 0.3083 & 19 & 0.4595 & 48 \\
\midrule
Batch 4 & 1/79 & \textbf{2025-DMIS-KU-4} & 0.9487 & 3 & 0.6136 & 2 & 0.6328 & 3 \\
& 40/79 & Fleming-1 & 0.9097 & 4 & 0.4697 & 10 & 0.4697 & 42 \\
& 58/79 & Fleming-5 & \textbf{0.9532} & \textbf{2} & 0.3311 & 16 & 0.3743 & 55 \\
& 63/79 & Fleming-4 & 0.9023 & 5 & 0.3250 & 17 & 0.3743 & 55 \\
& 64/79 & Fleming-2 & 0.9023 & 5 & 0.3250 & 17 & 0.3208 & 59 \\
& 65/79 & Fleming-3 & 0.8595 & 8 & 0.3250 & 17 & 0.3208 & 59 \\
\bottomrule
\end{tabular}
\label{tab:PhaseB_exact}
\end{table} 

\subsubsection{Ideal answer prediction}

Regarding the evaluation of the ideal answer for both Phase A+ and Phase B of Task 13b, we are currently waiting for the release of the scores manually assigned by the BioASQ experts, which are expected to be published shortly before the CLEF workshop in September. We note that all results for Task 13b remain provisional, as small corrections may still be applied by question curators prior to the workshop. 

\section{Conclusion and Future Work}
In this work, we presented a robust and extensible methodology for biomedical question answering within the BioASQ challenge framework. A key innovation in our methodology is the application and generalization of an LLM  ‘farming’ strategy, initially developed for Yes/No questions, to all exact question types. By systematically evaluating 13 state-of-the-art open LLMs and exhaustively testing all possible model combinations, we created optimized model farms for Yes/No, factoid, and list type questions. Our results show that combining multiple models improves performance in each case.
For Factoid questions, the best results came from combining six different LLMs. This pattern was consistent across both the BioASQ11 and BioASQ12 datasets, suggesting that the improvement wasn’t specific to the training data but rather due to the different strengths of each model working together. The top-performing combinations are shown in Table \ref{tab:best_llm_combinations}. For List questions, using too many models actually reduced performance. The best results came from small groups of about three models, for example, [DSQ8, L3.3, Qwen14], which outperformed all single models. 
For Yes/No questions, smaller combinations also worked best. Groups of three to four models, like the jury of Aya, Qwen32, Smaug, outperformed individual models.
In summary, combining LLMs can improve performance, but the optimal number of models depends on the question type.

Moving forward, we plan to expand our evaluation to include more state of the art open-source LLMs, incorporate more confidence scoring mechanisms across model outputs to better weigh and reconcile conflicting answers and release our question answering system
to support reproducibility and foster collaboration in the open LLM community.


\begin{acknowledgments}
We thank the anonymous reviewers for their valuable questions and comments, which helped us better showcase the significance of our work. We express our gratitude to the BioASQ challenge organizers for organizing the event and offering continuous support.
The GPU computations were executed on two servers acquired as part of project ID 16624, titled "Creation - Expansion - Upgrading of the Infrastructures of research centers supervised by the General Secretariat for Research and Innovation (GSRI)" with the code MIS 5161770. This project received funding under the "National Recovery and Resilience Plan Greece 2.0".
\end{acknowledgments}

\bibliography{sample-ceur}


\appendix
\section{Appendix}\label{Appendix}
In all these prompts, the \%s after QUESTION is replaced by the actual question, and the \%s after INFORMATION, TEXT or ABSTRACT is replaced with the collection of the related snippets or abstracts, concatenated and separated by a single blank.
\begin{tcolorbox}[enhanced,frame style ,
  opacityback=0.75,opacitybacktitle=0.25,
  colback=blue!35!green!13!white,colframe=blue!60!green!35!black!60,
  title=Yes/No Prompt ]
                Given only the following \textbf{INFORMATION} and \textbf{QUESTION}, answer the \textbf{QUESTION} only with "Yes" or "No". Think carefully.  \textbf{INFORMATION}: \%s \textbf{QUESTION}: \%s
\end{tcolorbox}

\begin{tcolorbox}[enhanced,frame style ,
  opacityback=0.75,opacitybacktitle=0.25,
  colback=blue!35!green!13!white,colframe=blue!60!green!35!black!60,
  title=List Prompt ]
                Answer the \textbf{QUESTION} using only the \textbf{TEXT} by only returning a list of entity names, numbers, or similar short expressions that are an answer to the question and are separated by commas. Only the list should be returned. If you do not know any answer return the word EMPTY. \textbf{TEXT}: \%s \textbf{QUESTION}: \%s
\end{tcolorbox}

\begin{tcolorbox}[enhanced,frame style ,
  opacityback=0.75,opacitybacktitle=0.25,
  colback=blue!35!green!13!white,colframe=blue!60!green!35!black!60,
  title=Factoid Prompt ]
                Answer the \textbf{QUESTION} using only the \textbf{TEXT} by only returning a list of entity names, numbers, or similar short expressions that are an answer to the question and are separated by commas,ordered by decreasing confidence. Only the list should be returned. If you do not know any answer return the word EMPTY. \textbf{TEXT}: \%s \textbf{QUESTION}: \%s
\end{tcolorbox}

\begin{tcolorbox}[enhanced,frame style ,
  opacityback=0.75,opacitybacktitle=0.25,
  colback=blue!35!green!13!white,colframe=blue!60!green!35!black!60,
  title=Summary Prompt ]
                \#\#\textbf{ABSTRACT}: \%s \#\#\textbf{QUESTION}: \%s \#\#\textbf{TASK}: Answer the \textbf{QUESTION} by returning a single paragraph sized text ideally summarizing only the most relevant information in the \textbf{ABSTRACT}.
\end{tcolorbox}

\begin{table}[h]
\centering
\caption{Best-performing LLM combinations by question type}
\label{tab:best_llm_combinations}
\begin{tabular}{|l|p{5cm}|c|p{5cm}|}
\hline
\textbf{Question Type} & \textbf{Best Model(s)} & \textbf{\# of LLMs} & \textbf{Notes} \\
\hline
Factoid & DeepSeek-R1-Distill-Llama-70B, LLaMA 3.3, Qwen3-14B, Qwen3-32B, Reflection, Smaug & 6 & Larger combinations performed best due to complementary strengths \\
\hline
List &  DeepSeek-R1-Distill-Llama-70B, LLaMA 3.3, Qwen3-14B & 3 & Small groups outperformed individual models and larger combinations \\
\hline
Yes/No & Aya, Qwen3-32B, Smaug & 3 & Small combinations (3–4 models) yielded the highest accuracy \\
\hline
\end{tabular}
\end{table}

\end{document}